\pdfoutput=1
\documentclass{article}
\usepackage{stywhispers,amsmath,amssymb,graphicx,booktabs}
\usepackage{hyperref}
\usepackage{orcidlink}
\hypersetup{hidelinks}
\providecommand{\IEEEmembership}[1]{#1}

\def\x{{\mathbf x}}
\def\d{{\mathbf d}}
\def\m{{\boldsymbol \mu}}
\def\C{{\mathbf C}}

\title{HUMAN-IN-THE-LOOP SIGNATURE BOOTSTRAPPING FOR UAV HYPERSPECTRAL PFM-1 MINE DETECTION}

\name{\begin{tabular}{@{}c@{}}
Sagar Lekhak\textsuperscript{1\textdagger}\orcidlink{0009-0009-7896-6167}~\IEEEmembership{Graduate Student~Member, IEEE},
Prasanna Reddy Pulakurthi\textsuperscript{1\textdagger}\orcidlink{0000-0003-0486-0756}, \\
Emmett J. Ientilucci\textsuperscript{1}\orcidlink{0000-0002-3643-8245}~\IEEEmembership{Sr.~Member, IEEE}
\end{tabular}%
\thanks{Corresponding author: Sagar Lekhak (Email: sl3088@rit.edu).}%
\thanks{\textsuperscript{\textdagger} These authors contributed equally to this work.}}
\address{\textsuperscript{1} Rochester Institute of Technology, Rochester, NY 14623, USA}

\begin{document}
\maketitle

\begin{abstract}
Hyperspectral imaging (HSI) is useful for material discrimination, but
operational mine screening also depends on how many false alarms must be
inspected before targets are found. This paper studies PFM-1 landmine
detection in unmanned aerial vehicle (UAV) visible and near-infrared (VNIR)
HSI using spectral angle mapper (SAM), matched filter (MF), adaptive coherence
estimator (ACE), and constrained energy minimization (CEM). We compare a
ground-measured SVC signature, a fully informed in-scene core-pixel signature,
and a simulated human-in-the-loop signature bootstrap.
Besides receiver operating characteristic area under the curve and average
precision, we report target-discovery curves and spatial candidate-review
counts. Full-review bootstrapping reaches the fully informed in-scene
signature case after all seven target regions are verified, but the required
inspection effort varies strongly: ACE confirms all regions in two rounds and
nine candidate inspections, whereas the SAM variants need thousands of
candidate reviews for their final target locations. Code is available at
\url{https://github.com/SagarLekhak/IEEE_WHISPERS_2026_UAV_HSI_PFM1}.
\end{abstract}


\section{Introduction}
\label{sec:intro}

HSI has been investigated for stand-off landmine detection because calibrated
wavelength-resolved spectra can help separate surface materials
\cite{makki2017survey}. UAVs make this sensing mode more practical for
close-range surveys while reducing direct personnel exposure in contaminated
areas \cite{bajic2017uav,baur2021drones}. Recent UAV VNIR datasets now support
controlled studies of PFM-1 detection with calibrated imagery, pixel labels,
and field spectroradiometer measurements
\cite{lekhak2025dataset,lekhak2026benchmark}.

SAM, MF, ACE, and CEM are widely used HSI target detectors when only a target
spectrum is available
\cite{manolakis2002detection,nasrabadi2014overview}. Their performance,
however, depends on how well that spectrum matches the target as imaged in a
specific scene. Illumination, view geometry, surface condition, sensor
response, reflectance retrieval, noise, and target--background mixing can
shift the apparent signature. Prior work has addressed this mismatch with
in-scene modeling, adaptation, and operator-feedback strategies
\cite{axelsson2016inscene,wang2014iterative,gerster2025tasr,schaum2009operator}.
Machine-learning models are outside the scope of this experiment because the
intended operational screening setting assumes little to no scene-specific
labeled training data; these classical detectors require only a target
signature and background statistics.

This paper therefore does not claim a new detector or a new adaptation theory.
Instead, it addresses a focused operational question in UAV-based PFM-1
screening: to what extent does target-signature selection influence detection
performance, and how effectively can operator verification of
detector-proposed locations recover the advantages of using an in-scene target
signature? These questions are investigated in a controlled experimental
setting designed to emulate key aspects of a real-world screening workflow. We
perform a retrospective within-scene comparison on seven physical target
regions, holding the image, detectors, and background statistics fixed while
changing only the signature source. To make the evaluation operator-facing, we
report both pixel-level metrics and the false-alarm burden before each target
region is discovered.

\section{Data and Methods}
\label{sec:methods}

\subsection{UAV hyperspectral scene}
\label{ssec:data}

The experiment uses a low-altitude PFM-1 subset from a UAV VNIR benchmark
dataset \cite{lekhak2025dataset,lekhak2026benchmark}. The analyzed image
contains $1705\times3461$ pixels and 272 retained bands spanning approximately
400--1000~nm. A binary pixel mask identifies visible PFM-1 target support. The
mask contains 248 positive pixels; all remaining 5,900,757 pixels are treated
as background. The target fraction is therefore approximately 0.0042\%, making
the scene strongly imbalanced.

\subsection{Notation and target detectors}
\label{ssec:detectors}

Let $\x_i\in\mathbb{R}^{B}$ denote the spectrum at pixel $i$, where $B$ is the
number of HSI bands, and let $\d$ denote the target signature resampled to the
image wavelengths. The scene mean and covariance are denoted by $\m$ and
$\C$, with inverse covariance estimated using a regularized pseudo-inverse.
For centered detectors, define $\tilde{\x}_i=\x_i-\m$ and
$\tilde{\d}=\d-\m$. Standard SAM, MF, ACE, and CEM formulations are given in
\cite{manolakis2002detection,nasrabadi2014overview}; the UAV PFM-1
implementation context follows \cite{lekhak2025dataset,lekhak2026benchmark}.
Here we summarize only the choices needed to interpret the experiments.

All detector outputs are score maps in which larger values indicate stronger
target evidence. The conventional SAM baseline is the uncentered spectral
cosine between $\x_i$ and $\d$, which is monotonic with the negative SAM angle
and therefore gives the same ranking as the usual angle with reversed sign. We
also include a mean-centered SAM variant because centering can change the
ranking of small target pixels in cluttered UAV imagery; this is reported
separately from classical uncentered SAM. MF and ACE use mean-centered pixel
and target vectors with the regularized inverse covariance. CEM is evaluated
as the standardized-background variant implemented in the experiments, using
band-standardized pixels and the corresponding inverse correlation matrix.

\subsection{Region-level discovery evaluation}
\label{ssec:regions}

Pixel-level metrics can be difficult to interpret in strongly imbalanced
problems because ROC-AUC may remain high even when the highest-ranked positive
predictions include many false alarms, while precision--recall measures are
more sensitive to the positive-class rarity \cite{davis2006pr,saito2015pr}.
For demining, a second issue is that the operational unit is a physical target
location, not an isolated pixel. We therefore define a target region as the
group of labeled target pixels corresponding to one visible PFM-1 location.
Raw connected-component labeling produces nine fragments in the binary
ground-truth mask; nearby fragments were merged into seven target regions for
discovery analysis.

For each detector, pixels are sorted from highest to lowest score. For each
target region, we record the rank of the first target pixel and the number of
non-target pixels ranked above it. We also plot a cumulative target-discovery
curve, defined as the number of target regions discovered as accumulated false
alarms increase. This pixel-ranking view is complemented by a spatial
candidate-review analysis, where an operator inspects a finite number of
non-maximum-suppressed candidate locations rather than individual pixels.

\subsection{Signature cases and bootstrap protocol}
\label{ssec:bootstrap}

\begin{figure}[!t]
\centering
\includegraphics[width=\linewidth]{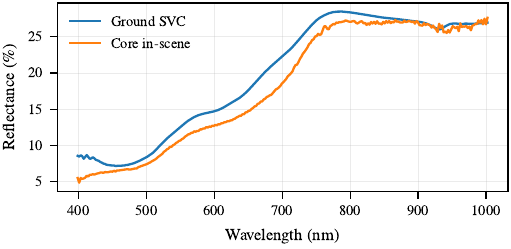}
\caption{Ground-measured SVC and ground-truth-derived core in-scene PFM-1
signatures after resampling to the UAV HSI wavelengths.}
\label{fig:sig}
\end{figure}

Three signature settings were evaluated. The external-signature case uses the
provided SVC spectroradiometer PFM-1 spectrum, linearly interpolated to the
272 HSI wavelengths. Although both the SVC signature and the UAV image are
expressed in reflectance, they come from different instruments, spatial
supports, and acquisition geometries. The core in-scene case uses the mean
spectrum of central target pixels from all seven target regions. This is a
fully informed reference, not a field-available input or a strict mathematical
upper bound, because averaging target pixels can suppress variability or favor
one detector. The bootstrap case starts from the SVC signature and updates the
signature only after detector-proposed locations are verified.
Fig.~\ref{fig:sig} compares the external and in-scene signatures used in these
experiments and illustrates the spectral mismatch that motivates the bootstrap
analysis.

For detector $a$ at review round $k$, let $\d_a^{(k)}$ be the current target
signature, $S_a^{(k)}$ its score map, $\mathcal{Q}_a^{(k)}$ the candidate set,
$\mathcal{T}_a^{(k)}$ the confirmed target regions, and
$\mathcal{B}_a^{(k)}$ the pixels covered by previous inspections. Let
$\mathcal{I}_{\eta}(q)$ denote the inspection neighborhood of radius $\eta$
around candidate location $q$. The bootstrap initializes $\d_a^{(1)}$ with the
external SVC signature, $\mathcal{T}_a^{(0)}=\emptyset$, and
$\mathcal{B}_a^{(0)}=\emptyset$. In each round,
$S_a^{(k)}(p)=g_a(\x_p;\d_a^{(k)},\m,\C)$ is recomputed from the current
signature, where $g_a$ is the scoring rule for detector $a$. The detector then
proposes the top $N$ spatially distinct unreviewed locations. Verified targets
contribute central pixels to the next-round in-scene signature; if no target
is verified, the signature is retained. The candidate set
$\mathcal{Q}_a^{(k)}\subset\Omega$ is selected by non-maximum suppression as
\begin{equation}
\mathcal{Q}_a^{(k)}=\operatorname{locNMS}_{N,\rho}
(S_a^{(k)},\Omega\setminus\mathcal{B}_a^{(k-1)}),
\end{equation}
where $\operatorname{locNMS}_{N,\rho}$ returns the top $N$ pixel locations
after spatial suppression radius $\rho$, and $\Omega$ is the image domain.
Human verification is represented by $v(q)$, a simulated operator-labeling
function using the ground-truth labels:
\begin{equation}
v(q)=
\begin{cases}
r, & \text{if } \mathcal{I}_{\eta}(q) \text{ intersects target region } r,\\
0, & \text{otherwise.}
\end{cases}
\end{equation}
The reviewed set is updated by blocking each inspected neighborhood so that
the same location is not repeatedly proposed:
\begin{equation}
\mathcal{B}_a^{(k)}=\mathcal{B}_a^{(k-1)}
\cup \bigcup_{q\in\mathcal{Q}_a^{(k)}} \mathcal{I}_{\eta}(q).
\end{equation}
The confirmed target set is updated by
\begin{equation}
\mathcal{T}_a^{(k)}=\mathcal{T}_a^{(k-1)}
\cup \{v(q):q\in\mathcal{Q}_a^{(k)}, v(q)>0\}.
\end{equation}
If at least one target region has been confirmed, the next signature is the
mean of the core pixels from all confirmed regions:
\begin{equation}
\d_a^{(k+1)} =
\frac{1}{|\mathcal{P}(\mathcal{T}_a^{(k)})|}
\sum_{p\in\mathcal{P}(\mathcal{T}_a^{(k)})}\x_p.
\end{equation}
If $\mathcal{T}_a^{(k)}=\emptyset$, then $\d_a^{(k+1)}=\d_a^{(k)}$. Here
$\mathcal{P}(\cdot)$ denotes the selected core pixels. For each target region
$r$, let $\mathcal{R}_r$ be its positive mask pixels and let $\bar{p}_r$ be
their spatial centroid. The core set for a confirmed collection
$\mathcal{T}$ is
\begin{equation}
\mathcal{P}(\mathcal{T})=
\bigcup_{r\in\mathcal{T}}
\operatorname*{arg\,min}_{\mathcal{U}\subseteq\mathcal{R}_r,\,
|\mathcal{U}|=\lceil f|\mathcal{R}_r|\rceil}
\sum_{p\in\mathcal{U}}\|p-\bar{p}_r\|_2^2 ,
\end{equation}
with $f=0.5$. Thus, the central 50\% of each confirmed target region is used
for signature averaging. In this subset, target region areas range from 11 to
61 pixels; when all seven regions are confirmed, 126 core pixels contribute to
the final signature.

In the experiments, $N=6$, $\rho=25$ pixels, and $\eta=12$ pixels; $N=6$
sets a small operator-review batch before each update. The flight altitude was
approximately 20.62~m \cite{lekhak2025dataset}; using a
same-sensor reference of 0.0188~m/pixel GSD at 30~m and assuming linear
scaling with height gives an estimated 0.0129~m/pixel GSD. Thus, $\rho$
imposes about 0.32~m candidate separation and $\eta$ corresponds to about a
0.16~m inspection radius. Because a PFM-1 is approximately
$120\times61\times20$~mm \cite{gichd2025ukraineguide}, these heuristic radii
are on the scale of the mine footprint while still reducing redundant
same-round proposals and allowing modest localization error. Review rounds are
numbered from one in
Table~\ref{tab:bootstrap}. The process stops when all seven target regions
are confirmed or when a preassigned inspection budget is exhausted. In this
retrospective analysis, the cap was 2001 review rounds, or at most 12,006
candidate reviews per detector. All detectors reached all seven target regions
before this cap; the required review counts are reported in
Table~\ref{tab:bootstrap}.

All detectors were run over the full image. Detection maps were retained as
raw scores for metric computation and min-max normalized only for
visualization. Thus, the normalized images do not change ROC, PR, AP, or
rank-based discovery statistics.

\section{Results}
\label{sec:results}

\begin{table*}[!t]
\centering
\caption{Pixel-level performance for target-signature settings. Full-review
bootstrap reports final maps after all seven target regions were confirmed.}
\label{tab:metrics}
\begin{tabular*}{\textwidth}{@{\extracolsep{\fill}}lcccccc@{}}
\toprule
& \multicolumn{2}{c}{Ground SVC} & \multicolumn{2}{c}{Core in-scene} & \multicolumn{2}{c}{Full-review bootstrap} \\
\cmidrule(lr){2-3}\cmidrule(lr){4-5}\cmidrule(lr){6-7}
Detector & ROC-AUC & AP & ROC-AUC & AP & ROC-AUC & AP \\
\midrule
SAM centered   & 0.6179 & 0.0077 & 0.7552 & 0.0797 & 0.7552 & 0.0797 \\
SAM uncentered & 0.7176 & 0.1277 & 0.8281 & 0.0100 & 0.8281 & 0.0100 \\
MF             & 0.9940 & 0.0385 & 0.9949 & 0.1546 & 0.9949 & 0.1546 \\
ACE            & \textbf{0.9974} & 0.1485 & \textbf{0.9987} & \textbf{0.3138} & \textbf{0.9987} & \textbf{0.3138} \\
CEM            & 0.9951 & \textbf{0.2136} & 0.9951 & 0.2547 & 0.9951 & 0.2547 \\
\bottomrule
\end{tabular*}
\end{table*}

Table~\ref{tab:metrics} summarizes the pixel-level results. ROC-AUC is high
for MF, ACE, and CEM in all signature cases, but AP gives a clearer comparison
because only 248 of 5,901,005 pixels are target pixels.
The ground SVC signature gives the strongest initial AP for CEM, whereas ACE
shows the largest AP gain from in-scene refinement. The full-review bootstrap
matches the core in-scene case because its final signature is built after all
seven regions have been verified.

\begin{figure}[!t]
\centering
\includegraphics[width=\linewidth]{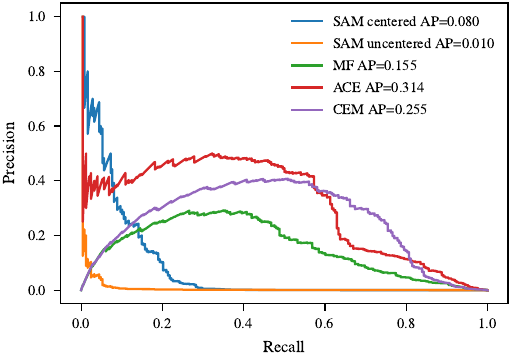}
\caption{Precision-recall curves for the full-review bootstrap score maps.}
\label{fig:pr}
\end{figure}

\begin{figure}[!t]
\centering
\includegraphics[width=\linewidth]{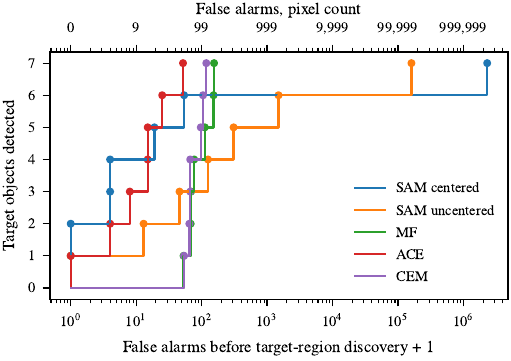}
\caption{Cumulative target discovery for the full-review bootstrap score
maps. Curves that rise earlier require fewer non-target inspections. The
$+1$ x-axis offset permits log-scale display when a target is discovered with
zero preceding false alarms.}
\label{fig:discovery}
\end{figure}

Figs.~\ref{fig:pr} and~\ref{fig:discovery} give complementary pixel- and
target-region views of the final bootstrap maps. Fig.~\ref{fig:pr} shows the
precision--recall behavior under the severe class imbalance, while
Fig.~\ref{fig:discovery} shows how quickly each detector reaches the seven
physical target regions as false alarms accumulate.

\begin{table*}[!t]
\centering
\caption{Candidate-review bootstrap discovery using six new spatial candidate
sites per detector per round. Milestones report the round and cumulative
inspected candidates (Cand.); the second column uses ``round: newly confirmed
region IDs''.}
\label{tab:bootstrap}
\begin{tabular*}{\textwidth}{@{\extracolsep{\fill}}lp{0.40\textwidth}rrrrrr@{}}
\toprule
& & \multicolumn{2}{c}{First target} & \multicolumn{2}{c}{Five targets} & \multicolumn{2}{c}{All targets} \\
\cmidrule(lr){3-4}\cmidrule(lr){5-6}\cmidrule(lr){7-8}
Detector & Confirmed target regions by round & Round & Cand. & Round & Cand. & Round & Cand. \\
\midrule
SAM centered   & 1: 3+5; 20: 2; 21: 1; 23: 7; 450: 4; 760: 6
               & \textbf{1} & 3 & 23 & 136 & 760 & 4558 \\
SAM uncentered & 1: 1+2+5; 2: 3+7; 6: 4; 483: 6
               & \textbf{1} & \textbf{1} & \textbf{2} & 9 & 483 & 2897 \\
MF             & 3: 1; 4: 3+4+5; 5: 2; 6: 7; 7: 6
               & 3 & 16 & 5 & 27 & 7 & 38 \\
ACE            & 1: 1+2+3+5; 2: 4+6+7
               & \textbf{1} & \textbf{1} & \textbf{2} & \textbf{7} & \textbf{2} & \textbf{9} \\
CEM            & 2: 1+3+4+5; 3: 2; 4: 6+7
               & 2 & 7 & 3 & 17 & 4 & 22 \\
\bottomrule
\end{tabular*}
\end{table*}

ACE has the highest AP and reaches all seven regions earliest in the discovery
curve. MF and CEM retain strong ROC-AUC but lower AP because some background
pixels remain highly ranked. Table~\ref{tab:bootstrap} translates these
rankings into spatial review effort: ACE reaches all seven regions in round 2,
after nine inspections, compared with 22 for CEM and 38 for MF. The SAM
variants find several targets early but require 4558 and 2897 reviews,
respectively, for the last region, indicating poor late-rank ordering.

\begin{figure*}[!t]
\centering
\begingroup
\setlength{\tabcolsep}{2pt}
\renewcommand{\arraystretch}{0.9}
\newcommand{\mappanel}[2]{%
\begin{minipage}[t]{0.295\textwidth}
\centering
\includegraphics[width=\linewidth]{#1}\\[-0.15ex]
{\footnotesize #2}
\end{minipage}}
\begin{tabular}{@{}ccc@{}}
\mappanel{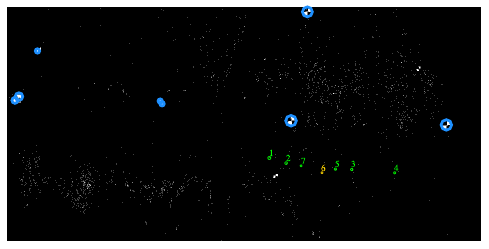}{(a) SAM centered} &
\mappanel{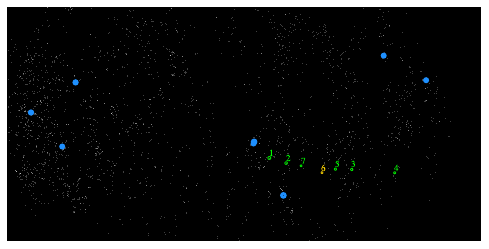}{(b) SAM uncentered} &
\mappanel{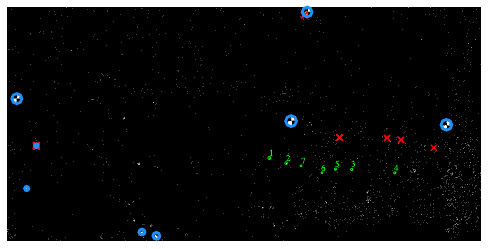}{(c) MF} \\[0.1ex]
\mappanel{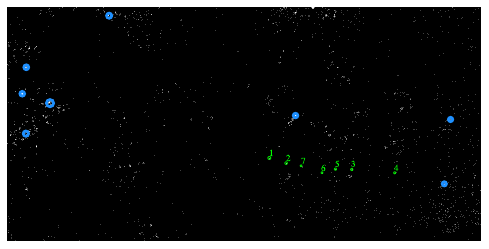}{(d) ACE} &
\mappanel{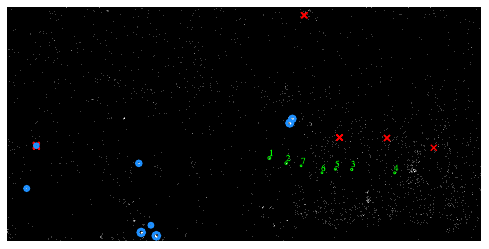}{(e) CEM} &
\end{tabular}
\endgroup
\caption{Full-review bootstrap detection maps thresholded at the 99.5th
score percentile for visualization. Green/gold circles mark detected/missed
target regions, red crosses false alarms before the first target hit, and blue
circles the largest binary false-alarm regions. Small numbers identify the
seven target-region IDs.}
\label{fig:maps}
\end{figure*}

Fig.~\ref{fig:maps} supports this interpretation spatially. At the common
99.5th-percentile display threshold, the SAM maps contain broad false-alarm
regions and miss one target region. MF and CEM concentrate more high scores
near the target line but still show isolated false alarms. ACE gives the most
compact high-score pattern, with all target regions represented and fewer
high-score background structures. Visual inspection indicates that several larger
false-alarm regions coincide with high-contrast calibration or georeferencing
objects deployed in the field, such as panels, ground control points, or
AeroPoints \cite{lekhak2025dataset}; in an operational workflow, known support
objects could be masked before candidate review.

\section{Discussion}
\label{sec:discussion}

The main implication is that HSI mine screening should be evaluated with
operator-facing metrics, not only pixel-wise separability. ROC-AUC can remain
strong even when high-ranking false positives precede some target regions.
Target-discovery curves and candidate-review counts directly address how much
non-target area must be inspected before each physical target location is
found.

The bootstrap experiment is retrospective, not a deployed autonomous system.
Ground truth only emulates human confirmation, and the core in-scene signature
is a fully informed reference rather than a field-available input. Even so,
verified in-scene target pixels can reduce signature mismatch, while stopping
after the first small candidate batch may reject detectors whose first target
appears deeper in the spatial ranking. Practical systems should track reviewed
locations and stop using an inspection budget, calibrated score threshold, or
patience rule. Future work should test adaptive candidate selection, diversity
constraints, and multi-detector proposal fusion under such budgeted review.

\section{Conclusion}
\label{sec:conclusion}

This case study evaluated SAM, MF, ACE, and CEM on UAV hyperspectral imagery
of PFM-1 targets under external, fully informed in-scene, and simulated
human-in-the-loop signature settings. Classical detectors remain useful
baselines, but their operational value depends on signature quality and early
false-alarm behavior. ACE was most efficient, confirming all seven target
locations in round 2 after nine reviewed candidates, while CEM and MF required
22 and 38 reviews. The SAM variants found several targets early but required
thousands of reviews for the final target locations. More broadly,
demining-oriented HSI studies should report target-discovery curves,
candidate-review counts, and false-alarm-before-detection values because
inspection burden is often as important as aggregate detection score.

\bibliographystyle{IEEEbib}
\bibliography{strings,refs}

\end{document}